\documentclass{article}

\usepackage{microtype}
\usepackage{graphicx}
\usepackage{subfigure}
\usepackage{booktabs} 

\usepackage{hyperref}

\usepackage[accepted]{icml2024}

\usepackage{amsmath}
\usepackage{amssymb}
\usepackage{mathtools}
\usepackage{amsthm}

\usepackage[capitalize,noabbrev]{cleveref}

\usepackage[textsize=tiny]{todonotes}

\usepackage{hyperref}
\usepackage{url}
\usepackage{siunitx}
\usepackage{subfigure}
\usepackage{multirow}
\usepackage{array}
\usepackage{booktabs}
\usepackage{overpic}
\usepackage{xspace}
\usepackage{tikz}
\usepackage{wrapfig}
\hypersetup{colorlinks, citecolor=green, linkcolor=red, urlcolor=blue}
\usepackage{arydshln}
\usepackage{pifont}
\usepackage{caption}
\usepackage{mathtools}
\usepackage{pythonhighlight}
\usepackage{enumitem}

\definecolor{TartOrange}{HTML}{ff2e35}
\definecolor{Orange}{HTML}{ff7825}
\definecolor{Mango}{HTML}{ffc013}
\definecolor{AppleGreen}{HTML}{7cb81b}
\definecolor{Blue}{HTML}{1173b0}
\definecolor{BdazzledBlue}{HTML}{2e58a5}
\definecolor{Purple}{HTML}{5b3590}
\definecolor{Sunglow}{HTML}{FFCA3A}
\definecolor{TableRow}{gray}{0.9}

\newcommand{\metricname}{\textsc{SkewSize\xspace}}
\newcommand{\Gap}{\textsc{Gap}}

\icmltitlerunning{Evaluating Model Bias Requires Characterizing its Mistakes}

\begin{document}

\twocolumn[
\icmltitle{Evaluating Model Bias Requires Characterizing its Mistakes}

\begin{icmlauthorlist}
\icmlauthor{Isabela Albuquerque}{gdm}
\icmlauthor{Jessica Schrouff}{gdm}
\icmlauthor{David Warde-Farley}{gdm}
\icmlauthor{Taylan Cemgil}{gdm}
\icmlauthor{Sven Gowal}{gdm}
\icmlauthor{Olivia Wiles}{gdm}
\end{icmlauthorlist}

\icmlaffiliation{gdm}{Google DeepMind}

\icmlcorrespondingauthor{Isabela Albuquerque}{isabelaa@google.com}

\icmlkeywords{Machine Learning, ICML}

\vskip 0.3in
]



\printAffiliationsAndNotice{} 

\begin{abstract}

The ability to properly benchmark model performance in the face of spurious correlations is important to both build better predictors and increase confidence that models are operating as intended.
We demonstrate that \emph{characterizing} (as opposed to simply quantifying) model mistakes across subgroups is pivotal to properly reflect model biases, which are ignored by standard metrics such as worst-group accuracy or accuracy gap.
Inspired by the hypothesis testing framework, we introduce \metricname, a principled and flexible metric that captures bias from mistakes in a model's predictions. It can be used in multi-class settings or generalised to the open vocabulary setting of generative models. \metricname~ is an aggregation of the \textit{effect size} of the interaction between two categorical variables: the spurious variable representing the bias attribute the model's prediction. We demonstrate the utility of \metricname~in multiple settings including: standard vision models trained on synthetic data, vision models trained on \textsc{ImageNet}, and large scale vision-and-language models from the \textsc{Blip-2} family. In each case, the proposed \metricname~is able to highlight biases not captured by other metrics, while also providing insights on the impact of recently proposed techniques, such as instruction tuning.
\end{abstract}
\section{Introduction}

Machine learning systems can capture unintended biases~\citep{dixon2018measuring} by relying on correlations in their training data that may be spurious (i.e.~a finite sample artifact), undesirable and/or that might vary across environments. 
Models of all scales are vulnerable to this failure mode, including recent, large-scale models~\citep{weidinger2022taxonomy,birhane2023hate,luccioni2023stable,solaiman2023evaluating}.
To evaluate unintended biases in model outputs, existing metrics divide the population (or test set) into subgroups (based on demographic characteristics, how well-represented in the dataset each group is, or another characteristic of significance) and aggregate the e.g.\ correct and incorrect outputs across those subgroups as in \cite{sagawa2019distributionally}.
However, existing metrics consider as equivalent all responses deemed to be incorrect, obscuring important information regarding a model's bias characteristics, especially in the context of large or intractable output spaces.
\begin{figure*}
    \centering
    \includegraphics[width=0.89\linewidth]{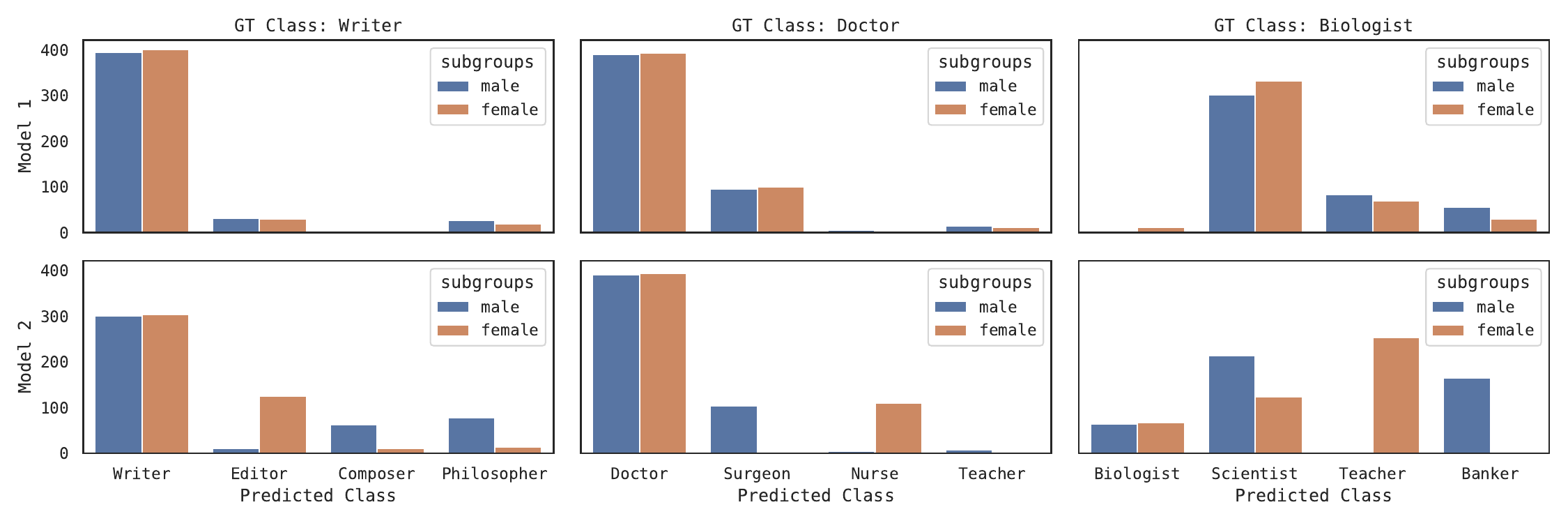}
    \scriptsize
    \begin{tabular}{lccc|ccc|ccc|ccc} \toprule
         & \multicolumn{3}{c}{\bf Writer} & \multicolumn{3}{c}{\bf Doctor} & \multicolumn{3}{c}{\bf Biologist} & \multicolumn{3}{c}{\bf All} \\
         & M1 & M2 & $\Delta\downarrow$ & M1 & M2 & $\Delta\downarrow$ & M1 & M2 & $\Delta\downarrow$ & M1 & M2 & $\Delta\downarrow$ \\ \midrule
         Acc  & 0.88 & 0.67 & -0.21 & 0.77 & 0.77 & 0.00 & 0.02 & 0.15 & 0.13 & 0.56 & 0.53 & -0.03 \\
         WG   & 0.87 & 0.66 & -0.21 & 0.77 & 0.77 & 0.00 & 0.01 & 0.14 & 0.14 & 0.55 & 0.53 & -0.02 \\
         \Gap & 0.01 & 0.00 & 0.00 & 0.00 & 0.00 & 0.00 & 0.01 & 0.00 & -0.01 & 0.01 & 0.00 & 0.00 \\
         - \metricname~& -0.07 & -0.45 & {\bf -0.38} & -0.04 & -0.45 & {\bf -0.40} & -0.12 & -0.71 & {\bf -0.58} & -0.44 & -0.71 & {\bf -0.27} \\ \bottomrule
    \end{tabular}
    \caption{{\bf Standard metrics fail to capture biases within a model.} We plot the prediction counts for two models given three ground-truth classes (\textit{Writer}, \textit{Doctor}, \textit{Biologist}). \textsc{Model 1 (M1)} displays similar distributions of errors for both subgroups whereas \textsc{Model 2 (M2)} displays ``stereotypical'' errors (e.g. mispredicting female \textit{Doctor}s for \textit{Nurse}s). In the table, we report accuracy (Acc), worst group accuracy (WG), \Gap~and their difference ($\Delta$) between \textsc{M1} and \textsc{M2}.  Only our approach (\metricname) captures the bias in all settings.}
    \label{fig:motivatingexample}
\end{figure*}

\textbf{Motivating example.}
Consider the synthetic setup in~\autoref{fig:motivatingexample} which compares two image classification models: \textsc{Model 1} and \textsc{Model 2}. These models predict occupation, with different distributions of outputs across two mutually exclusive\footnote{Assumed to be mutually exclusive for the limited purpose of this illustrative example. We recognize that reality is richer and more nuanced than this binary categorization. See \emph{Impact Statement} section.} subgroups (\textit{male} and \textit{female}). Following prior work, we first compute model accuracy in each subgroup~\citep[e.g.\ ][]{chowdhery2022palm}, worst group accuracy \citep[i.e. minimum accuracy across groups, ][]{sagawa2019distributionally} and \Gap~ \citep[the difference between subgroup accuracy and overall accuracy,][]{zhang2022contrastive} across the following three ground-truth classes:
\vspace{-0.4cm}
\setlist{nolistsep}
\begin{itemize}
\itemsep0.2em
\item \textsc{Writer}: \textsc{Model 2}'s accuracy is lower than that of \textsc{Model 1}; a bias in \textsc{Model 2}'s predictions is evident in women being misclassified as \textit{Editor}s and men being misclassified as \textit{Composer}s and \textit{Philosopher}s. Accuracy and Worst group accuracy degrade as expected for the more biased model, whereas \Gap~does not.
\item \textsc{Doctor}: Accuracy is the same for \textsc{Model 1} and \textsc{Model 2} but a bias is evident in  \textsc{Model 2}'s predictions, with women being misclassified as \textit{Nurse}s, and men being misclassified as \textit{Surgeon}s. Traditional accuracy-based metrics do not capture this bias.
\item \textsc{Biologist}: Accuracy is higher for \textsc{Model 2} than \textsc{Model 1}, but a bias is evident in  \textsc{Model 2}'s predictions, with women being misclassified as \textit{Teacher}s and men as \textit{Scientist}s or \textit{Banker}s. Counterintuitively, the standard metrics improve or stay the same.
\item \textit{All}: Aggregating across classes, we can see that the standard metrics either improve in \textsc{Model 2} relative to \textsc{Model 1} or do not change.
\end{itemize}

\vspace{-0.3cm}
In light of this example, we see that regardless of how the performance of a model \textit{in terms of accuracy} varies across subgroups, bias may also arise from systematic errors in incorrect predictions.
Importantly, previously proposed metrics do not surface such bias and give the misguided impression that the model's predictions do not exhibit bias. 
To measure this type of bias, we introduce \textsc{\metricname}, which considers \textit{how different} the distribution of predictions are across subgroups. In our motivating example, \textsc{\metricname} is able to capture the different types of biases.

We propose to formulate the problem of estimating bias for classification models through the lens of hypothesis testing.
We draw inspiration from tests of association between the confounding, spurious factor (e.g.\ gender) and the model's prediction, and propose to re-purpose a measure of \emph{effect size} for such tests. We compute effect sizes of this association for each ground-truth class: for instance, given images of doctors, we can estimate the effect size corresponding to the association between gender and predicted occupation. As shown in Section ~\ref{sec:experiments_dsprites}, this approach yields a fine-grained and interpretable assessment of model bias, exposing the most affected classes, as opposed to accuracy-based or fairness metrics.
Finally, we propose to aggregate effect sizes across classes using a measure of the skewness of the effect size distribution per class to arrive at a scalar metric which can be used to compare different models. We validate the metric and investigate its utility in three settings: {\em(1) Controlled data generation:} First, we investigate the utility of the metric, relative to existing ones, by validating it on a controlled setting where spurious correlations affecting single classes are induced, and demonstrate it not only reliably captures bias but also correctly exposes the affected classes.

{\em(2) Multi-class classification:} Next, we demonstrate how our metric can be used to identify previously unidentified cases of systematic bias arising from model mispredictions using \textsc{DomainNet} and \textsc{ImageNet}~\citep{deng2009imagenet,peng2019moment}.
{\em(3) Open-ended prediction:} Finally, we analyse large scale vision-and-language models (VLMs) (\textsc{Blip-2}, \cite{li2023blip}) that have an intractable\footnote{This refers to the setting where the label space is given by all the possible outputs of a language model.} output space using our metric in two separate settings: gender vs.~occupation and gender vs.~practiced sport.
In settings (2) and (3), we find that no datasets exist that would allow for computing bias metrics among subgroups with statistical significance. 
As a result, we create synthetic datasets in order to run our evaluation at scale. Our main contributions are summarized as follows:
\vspace{-0.35cm}
\begin{enumerate}[leftmargin=*]
\itemsep0.05em 
    \item We demonstrate limitations of current metrics for quantifying bias, specifically that they fail to capture bias manifested in {\em how the model makes mistakes}.
    \item  We propose \metricname, a metric for evaluating bias in discriminative models inspired by hypothesis tests of contingency tables.
    \item We use \metricname~to evaluate model bias at scale in a variety of domains and identify biases arising in the models' errors. We further show how \metricname\xspace can be used with synthetic data to evaluate bias in VLMs. 
\end{enumerate}

\section{Method}
\label{sec:method}

\subsection{Background}\label{sec:backround}
\textbf{Notation.} We consider a discriminative model $f_{\theta} : \mathcal{X} \mapsto \mathcal{Y}$ with parameters $\theta$, where $\mathcal{X}$ is the set of inputs (e.g.\ images) and $\mathcal{Y}$ is the label set.
We also assume that input $x \in \mathcal{X}$ with label $y \in \mathcal{Y}$ is drawn from an underlying distribution $p(x|z,y)$, where $z$ is a discrete latent variable $z \in \mathcal{Z}$ that represents a factor of variation that affects the data generating process. In the context of this work, $z$ is referred to as the \textit{bias} variable and assumed to systematically affect how well the model $f_{\theta}$ is able to predict $y$ from $x$. Our goal is then estimating to what extent the predictions are affected by $z$.

\textbf{Metrics for output disparity across subgroups.} Previous work on evaluating performance disparity across subgroups has mostly considered metrics such as accuracy \citep{zhang2022contrastive,alvi2018turning,li2022discover}, worst group accuracy \citep{zhang2022contrastive,koh2021wilds}, gap between average and worst group accuracy \citep[referred to as \Gap, ][]{zhang2022contrastive}. These metrics focus on the true positive rate and do not identify biases in the distribution of prediction errors. We compute these metrics throughout the work, for comparison with our approach. 

Alternatively, fairness criteria can be formulated as independence desiderata~\citep{Barocas2019-kj}, with metrics classified as `independence' criteria if $f_{\theta}(x) \perp z$, `separation' if $f_{\theta}(x) \perp z | y$ and `sufficiency' if $ y \perp z | f_{\theta}(x)$. In practice, these criteria are operationalized using different metrics. For the independence criterion, demographic parity \citep[DP]{dwork2012} is commonly used. These metrics have been recently extended for use in the multiclass setup \citep[e.g.][]{alabdulmohsin2022a,pagano2023bias,putzel2022blackbox,rouzot2022learning}. In this case, metrics are typically computed by binarizing each class \citep[e.g.][]{alabdulmohsin2022a,pagano2023bias} and aggregating fairness scores across classes using their maximum (i.e. worst case scenario), or average (c.f. Appendix ~\ref{sec:appendix_fairness_metrics}). Given a full confusion matrix, equality of odds (EO)~\citep{hardt2016}, and potentially DP, would capture differences in the distributions of model errors. However, the detected bias would be surfaced in the scores of the confused classes rather than associated with the class of interest. In our motivating example, EO comparing \textsc{male} and \textsc{female} examples in the \textsc{Doctor} class would be close to 0, but larger for the \textsc{Surgeon} and \textsc{Nurse} classes. In an intractable output space, a full confusion matrix may be unavailable, and EO and DP would be limited in their ability to highlight differences in the distribution of model errors. In this work, we compute EO and DP as per \citet{alabdulmohsin2022a} when a full confusion matrix is available.

\subsection{Estimating distributional bias for categorical distributions}
Let $\mathcal{Z}=\{A, B\}$ and $\mathcal{Y}$ be a discrete set. We further consider that the parametric model $f_{\theta}(x)$ defines a conditional distribution $q(y|x;\theta)$ for each $x\in\mathcal{X}$. For a fixed value of $y' \in \mathcal{Y}$, distributional bias should account for systematic differences in the outcomes of $f_{\theta}(x)$ across different subgroups, i.e. when $x$ is sampled from $p(x|y,z=A)$ versus $p(x|y,z=B)$. More formally, in Equation~\ref{eq:dist_bias_divergence}, we define distributional bias as a comparison between \textit{induced families of distributions} defined by $f_{\theta}(x)$ when $x \sim p(x|y=y',z=A)$ versus when $x \sim p(x|y=y',z=B)$: 
\begin{equation}
     \mathcal{H}(Q_A(y|x;\theta) || Q_B(y|x;\theta)),
    \label{eq:dist_bias_divergence}
\end{equation}
where $Q_A(y|x;\theta)$ and $Q_B(y|x;\theta)$ denote the family of distributions obtained when the bias variable assumes each of its possible values, i.e. $z=A$ and $z=B$, respectively. $\mathcal{H}(\cdot||\cdot)$ is an operator that accounts for a notion of similarity between the two distributions. Depending on the nature of $Q$, $\mathcal{H}$ can assume different forms. Also, notice that $\mathcal{H}$ operator is not limited to binary attributes and can be instantiated by approaches to compare families of distributions.

As we focus on classification tasks, $f_{\theta}(x)$ parameterizes families of categorical distributions. We can thus formulate the comparison between $Q_A$ and $Q_B$ as \textit{estimating the effect size}, i.e. the practical relevance of the association between the bias variable $z$ and predictions $y'\sim q(y|x,z)$. In this framework, the similarity between $Q_A$ and $Q_B$ can be seen as a measure of association between two \textit{categorical} variables, $z$ and $y'$. Given that, effect size can then be estimated via the Cram\'er's V statistic \citep{cramer1946mathematical}, defined as:
\begin{equation}
\nu = \sqrt{\frac{\chi^2}{N \cdot DF}},
\label{eq:effect_size}
\end{equation}
where $N$ is the sample size, $DF$ is the number of degrees of freedom,
and $\chi^2$ represents the test statistic from the corresponding Pearson's chi-squared independence test. Cram\'er’s V is bounded between 0 and 1, with 1 indicating a perfect association between both variables, i.e.\ the predictions are extremely sensitive to changes in the value of the bias variable. In order to compute the value of $\chi^2$, the counts of predictions must be arranged in a \textit{contingency table} of size $M=|\mathcal{Z}|\cdot|\mathcal{Y}|$. Contingency tables show the frequency distribution of the variables under consideration, therefore, for a given class $y'$, each entry should correspond to the frequency with each predicted class was observed per subgroup in the data. The $\chi^2$ statistic for such observations accounts for the discrepancy between observed and expected frequencies in the table and is defined as:
\begin{equation}
    \chi^2 = \sum_{k=1}^{M}\frac{(o_k-e_k)^2}{e_k},
\label{eq:chi_square}    
\end{equation}
where $o_k$ refers to the observed value of the $k$-th entry in the contingency table, and $e_k$ refers to the expected value of this table entry under the assumption of independence between the bias variable and the prediction. In Appendix \ref{sec:appendix_mev} we propose and empirically validate an strategy to control for noise in the predictions when computing the quantity in Eq. \ref{eq:chi_square} as the size of output space $|\mathcal{Y}|$ increases.

\subsection{Aggregating the Effect Size}
The effect size based approach to measure distributional bias evaluates model predictions on a per-class basis. In order to obtain a single, scalar summary metric which can be used to compare multiple models, we must consider how to aggregate the estimated effect sizes for all classes. The ideal metric should be able to simultaneously satisfy the following two conditions: (i) indicate an overall weaker bias when the distribution of effect size values per class is centered around zero with infrequent higher values (as classes for which the model is strongly affected by bias are rare), (ii) distinguish models weakly exhibiting bias from models where, for a considerable fraction of classes, the predictions exhibit high degrees of association with the bias variable, (i.e., the distribution of effect size values is long-tailed and skewed towards the left).

Given the aforementioned desiderata, we  propose to aggregate the effect size values per class using the Fisher-Pearson coefficient of skewness, as it captures both how \emph{asymmetric} the distribution of estimated effect size values is as well as the \textit{direction} of the asymmetry. For estimated effect sizes $\{\nu_1, \nu_2, \ldots, \nu_{|\mathcal{Y}|}\}$ with empirical mean $\bar{\nu}$, the proposed metric \metricname~is computed as:
\begin{equation}
    \text{\metricname}= \frac{\sum_{i=1}^{|\mathcal{Y}|}(\nu_i-\bar{\nu})^3}{\left[\sum_{i=1}^{|\mathcal{Y}|}(\nu_i-\bar{\nu})^2\right]^{3/2}}.
    \label{eq:metric_def}
\end{equation}
In the Appendix we provide pseudocode for \metricname~(Alg.~\ref{alg:metric_computation}), a Python implementation, and a discussion on modulating the impact low count predictions might have.

\textbf{Flexibility of \metricname.}
As we are interested in surfacing biases in the distribution of model errors, our formulation can be related to a `separation' fairness criterion. However, this does not preclude the implementation of other criteria. For instance, sufficiency could be implemented by conditioning on the predicted class and using the ground-truth class as the independent variable in the $\chi^2$ test. Similarly, we can implement DP by using the model outputs as the independent variable. 

\metricname~can also be implemented considering other choices of statistics, as we show in Appendix \ref{sec:appendix_phi}. Here we choose Cram\'er's V as it is more general and applicable to contingency tables larger than 2x2. Finally, \metricname~can be computed based on logits, softmax scores, top-1 or top-$k$ predictions. Here, we focus on the separation formulation based on top-1 predictions in each class, in which case \metricname~ is also applicable to scenarios where this is the only information about the model’s output which is available to the user \citep{achiam2023gpt}.

\begin{table*}[!ht]
\centering
\resizebox{\textwidth}{!}{
\begin{tabular}{cccc|ccc|ccc||ccc}
\toprule
        & \multicolumn{3}{c|}{Accuracy-based} & \multicolumn{3}{c|}{DP$(\downarrow)$} & \multicolumn{3}{c||}{EO $(\downarrow)$} & \multicolumn{3}{c}{\textbf{Effect size} $(\downarrow)$}   \\
Removed                & Accuracy $(\uparrow)$   & WG $(\uparrow)$    & \Gap~$(\downarrow)$      & Class 0   & Class 1   & Class 2 & Class 0   & Class 1   & Class 2 & Class 0   & Class 1  & Class 2   \\ \midrule 
Unbiased               & 0.998      & 0.996 & 0.002   & 0.004     & 0.004     & 0.001   & 0.001     & 0.001    & 0.001   & 0.006     & 0.020     & 0.024  \\      
\color{green}{Class 0} & 0.888      & 0.666 & 0.222   & 0.050     & 0.315     & 0.265   & 0.038     & 0.483    & 0.203   & 0.705     & 0.012     & 0.017\\
\color{green}{Class 1} & 0.891      & 0.653 & 0.238   & 0.303     & 0.013     & 0.289   & 0.448     & 0.009    & 0.220   & 0.012     & 0.703     & 0.011\\
\color{green}{Class 2} & 0.888      & 0.664 & 0.224   & 0.278     & 0.057     & 0.332   & 0.208     & 0.040    & 0.484   & 0.032     & 0.009     & 0.700\\ \bottomrule
\end{tabular}}
\caption{\textbf{Effect Size captures bias due to introduced spurious correlations on \textsc{dSprites}.} Metrics computed on the predictions by ResNet18s trained on 4 versions of \textsc{dSprites} show that Effect size is only non-negligible for the corresponding biased classes and indicates which is the class affected by spurious correlations.}
\label{tab:toy_data}
\end{table*}

\section{Experiments}\label{sec:experiments}
We first empirically demonstrate the effectiveness of \metricname~to measure biases in a controlled experiment with the dSprites \citep{dsprites17} dataset. We then proceed to show the usefulness of \metricname~ and how it is complementary to accuracy-based metrics in ubiquitous tasks such as classification with the \textsc{ImageNet} \citep{deng2009imagenet} dataset and the \textsc{DomainNet} benchmark (in Appendix~\ref{sec:appendix_domainnet}). We finalize the experimental validation of \metricname~ with its application to assess \textsc{VLM}s from the~\textsc{Blip-2} family and demonstrate it can be used to uncover biases and compare models even in cases where predicted classes do not necessarily appear as ground-truth in the evaluation dataset.

\subsection{Controlled setting: \textsc{dSprites} dataset}\label{sec:experiments_dsprites}
We begin with a setting where controlled levels and types of bias can induced. We use the \textsc{dSprites} dataset, which contains images of objects represented by different shapes, colors and at different positions. We leverage knowledge about the data generating process to introduce spurious correlations in the training data by excluding examples of a specific shape and color. As in this case the effect of relying on such spurious correlations is reflected by accuracy-based metrics, this allows us to validate that effect size estimation can be used as a strategy to capture biased predictions and provides information on which class was affected by the introduced spurious correlation.

\textbf{Setting.} We consider the task of predicting the object \textit{shape} and evaluate whether the model predictions are biased with respect to object \textit{color} under a regime of systematic training set manipulation. Using the terminology in Section \ref{sec:method}, the object color is the \textit{independent variable} (i.e.~the variable on which we intervene), and the predicted shape is the \textit{dependent variable} (i.e.~the variable we observe). 
\textbf{Effect Size Captures different types of bias.}
Given the \textsc{dSprites}' label set has three classes and with color attributes in $\{\texttt{Red}, \texttt{Green}, \texttt{Blue}\}$, we build versions of the training data that have different spurious correlations between color and label by removing all examples in the \textsc{Green} color from one of the classes. We then train a ResNet18~\citep{he2016deep} for 5k steps with each dataset and evaluate on held out data that has \emph{not} been manipulated to remove any examples. For each ground-truth class, we compute the effect size of the interaction between color and prediction as described in Section~\ref{sec:method}. For completeness, we verify that each model reaches nearly 100\% accuracy on a test set biased in a similar way. These experiments highlight the existence of bias in the model's predictions: accuracy should be nearly 100\% for all classes but the one that had examples removed, in which case errors mostly occur when the object is \textsc{Green}.

We present in Table~\ref{tab:toy_data} results for the three models and one with the same architecture but trained with unbiased data. We report accuracy-based metrics on an \textit{unbiased} test set, along with EO, DP, and per-class effect size (our approach). As intended, models trained with biased data had lower accuracy in comparison with the model trained with unbiased data, suggesting they indeed rely on the introduced spurious correlation. We observe that, for all models, effect sizes were strong only for the classes affected by the spurious correlation (i.e. the ones that had green instances removed at training), while remaining negligible for the other classes, confirming that the proposed approach indeed captures model biases and correctly provides per-class granularity. In contrast, EO and DP tend to distribute the effect of this bias across the confused classes, and do not readily indicate the origins of the confusion.

\textbf{Effect Size Captures different bias levels.} Following a similar set-up to previous experiment, we now induce different levels of the same spurious correlation by creating training datasets containing different number of examples from the combination of color and class. We created three datasets by removing instances from \textsc{Class 1} in the \textsc{Green} color so that only $\{5k, 2k, 0\}$ of such examples are left in the training data. We adopt the same architecture, training, and evaluation from the previous experiment.
Results are shown in Table \ref{tab:dsprites_bias_levels}, where, for reference, we also report the performance of the unbiased model. As expected, accuracy-based metrics decreased as the number of examples from the removed class, color increased, confirming the models are increasingly affected by the induced spurious correlation. We find that the effect size for the affected class presents a monotonic increasing relationship with bias strength, effect size for the unaffected classes remained negligible, confirming that the effect size captures different levels of bias and correctly indicates the affected class.

\begin{table}[t]
\resizebox{0.45\textwidth}{!}{
\centering
\begin{tabular}{ccc||ccc}
\toprule
              & \multicolumn{2}{c||}{Acc.-based} & \multicolumn{3}{c}{\textbf{Effect size} $(\downarrow)$}       \\ 
Bias strength & Acc. $(\uparrow)$              & WG $(\uparrow)$             & Class 0   & Class 1           & Class 2  \\ \midrule
Unbiased      & 0.998         & 0.996          & 0.006     & 0.020             & 0.024    \\
Mild          & 0.977         & 0.936          & 0.002     & \textbf{0.253}    & 0.017    \\
Medium        & 0.933         & 0.806          & 0.013     & \textbf{0.492}    & 0.032    \\
Strong        & 0.891         & 0.653          & 0.012     & \textbf{0.703}    & 0.011    \\ \bottomrule
\end{tabular}}
\caption{\textbf{Inducing varying bias strengths in models trained on \textsc{dSprites}}. The bias strength denotes the number of examples from Class 1 in the green color that were left in the respective versions of the training data. No Bias: full training set, Mild: 5k, Medium: 2k, Strong: 0.}
\label{tab:dsprites_bias_levels}
\end{table}

\subsection{Estimating distributional bias in multi-class classification: \textsc{ImageNet}}
\begin{figure*}[!ht]
\centering
\includegraphics[width=\linewidth]{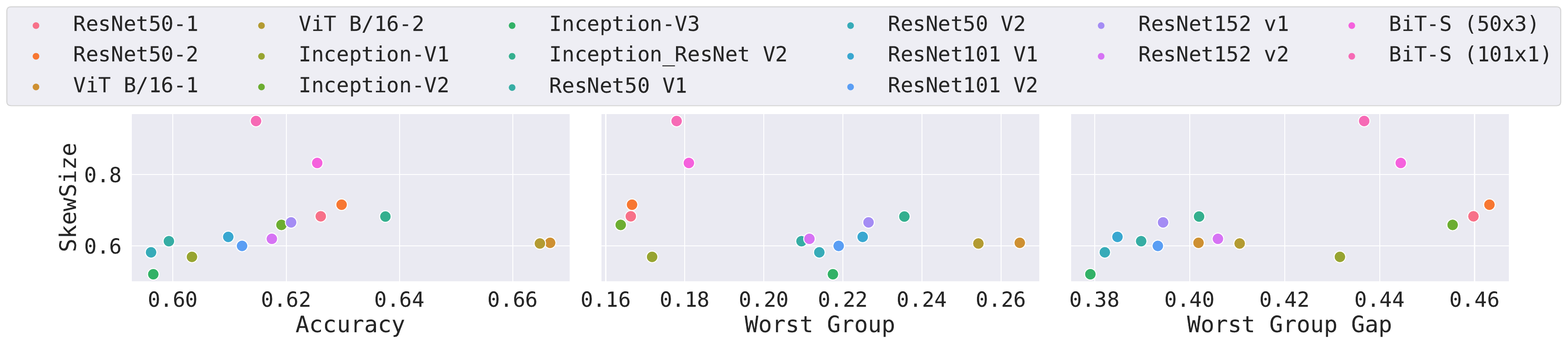}
\caption{\textbf{Comparing models trained on \textsc{ImageNet} across multiple metrics}. We plot \metricname~versus each accuracy-based metric for a variety of models. The results highlight that no accuracy-based metric presents a clear trend with respect to \metricname, demonstrating it captures aspects of performance not exposed by these other metrics. Moreover, models with similar performance according to accuracy-based metrics, such as both BiT-S models, can be discriminated by \metricname~.} 
\label{fig:imagenet_radar_sub1}
\end{figure*}

We have thus far demonstrated that \metricname~is capable of accounting for aspects of a model's behaviour that are not captured by accuracy-based bias metrics. We now showcase how \metricname~can be used to provide a more comprehensive evaluation of classifiers by distinguishing models that perform similarly in terms of accuracy, but turn out to display different levels of bias. 

\textbf{Models.} We consider models spanning four architectures: \textsc{ResNet50s} \citep{he2016deep}, \textsc{Vision Transformers (ViTs)} \citep{dosovitskiy2020image}, \textsc{Inception} \citep{szegedy2015going}, and \textsc{BiT} models \citep{kolesnikov2020big}. Architecture and training details are described in Appendix~\ref{app:imagenetmodels}.

\textbf{Data.} We consider a scenario where the background of an image corresponds to the bias variable to evaluate the \metricname~of each model. As no background annotations are available in the original \textsc{ImageNet}, we chose 200 classes from the original label set (specifically, those present in \textsc{TinyImageNet}~\citep{le2015tiny}) and generated a synthetic dataset containing images of each of the selected classes across 23 different background types (list obtained from \citet{vendrow2023dataset}) using \textsc{Stable Diffusion}~\citep{rombach2022high}.
We generate images using the prompt template \colorbox{blue!10}{\emph{A photo of a \{\textsc{class}\} \{\textsc{background}\}}}.
For instance, for the class \textsc{Salamander}, we used prompts such as \colorbox{blue!10}{\emph{A photo of a \textsc{salamander} \textsc{on the rocks}}}. We generate 200 images for each background-class pair. Note that these images are used only for evaluation, not training.

\textbf{Results.} In Figure \ref{fig:imagenet_radar_sub1} we compare models in terms of accuracy, worst group accuracy, worst group accuracy gap, and \metricname. The first aspect to observe is that, overall, no clear correlation between these metrics and \metricname: models with similar accuracy may present considerable disparities in how biased they are as demonstrated by the differences in \metricname~values. Specifically, we highlight that although models such as \textsc{BiT-S 50x3} and \textsc{101x1} present similar performance as per all considered accuracy-based metrics, they can be further discriminated by \metricname~as \textsc{BiT-S (101x1)} achieved higher a value for this metric. 

\begin{figure}[h]
\includegraphics[width=0.49\textwidth]{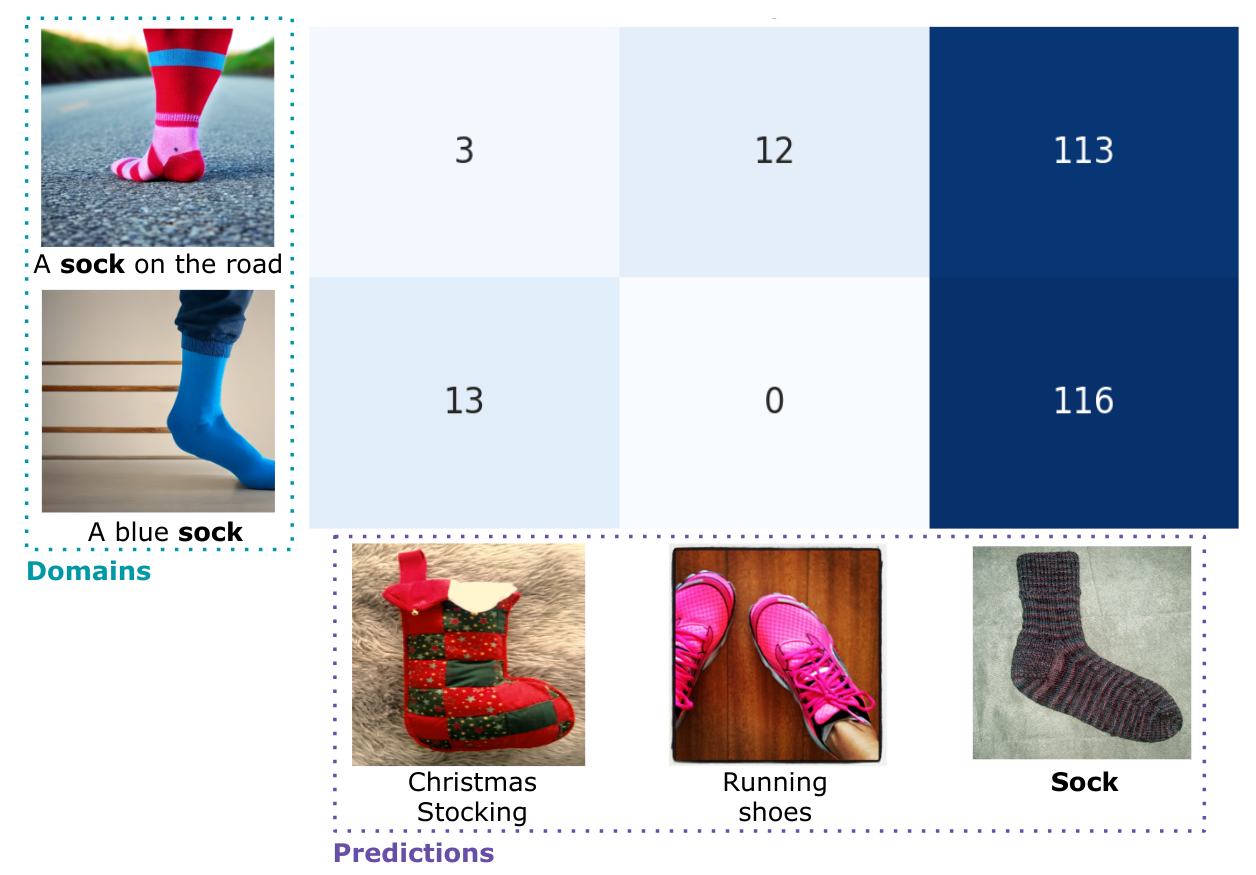} 
\caption{\textbf{Bias exposed by \metricname.} Both domains for the \textsc{socks} class have similar accuracy, but a mismatch in errors indicates the model relies on spurious features of background/color.}
\label{fig:socks_vit1}
\end{figure}

\textbf{Uncovering spurious correlations with \metricname.} We now examine specific cases of systematic bias uncovered by \metricname. We identify examples by investigating classes where the model is both accurate and the effect size for the association between background and the model's prediction is high. In Figure~\ref{fig:socks_vit1}, we show the top-3 predictions by the \textsc{ViT B/16-1} for $\textsc{socks}$ in subgroups corresponding to \colorbox{blue!10}{\emph{A photo of a \textsc{sock} \textsc{on the road}}} and \colorbox{blue!10}{\emph{A photo of a \textsc{blue} \textsc{sock}}}. Both sub-groups/domains present similar measured accuracy, in which case metrics such as worst group accuracy and \textsc{Gap} would be ineffective to capture bias that can be observed in the misclassified cases. This disparity in the distribution most frequent errors for each subgroup is in fact captured by \metricname~and suggest that the evaluated model may incorrectly associate an \textsc{on the road} background with the class \textsc{running shoes}, even when the true object of interest is absent.
\subsection{Comparing VLMs for multi-class classification across model size}

We now consider the case where the output space is intractable and obtaining data to evaluate the model is challenging. We study the \textsc{Blip-2} model family for (binary) gender bias when predicting occupation or practiced sport. 

\textbf{Data.} Apart from Visogender \citep{hall2023visogender}, with only 500 instances, there are no real-world datasets available for evaluating gender biases on VLMs. Therefore, to investigate the utility of \metricname\xspace in the evaluation of VLMs, we gather synthetic data with templates constructed as follows.
For the occupation task, we use templates of the form \colorbox{blue!10}{\emph{A \{\textsc{gender}\} \{\textsc{occupation}\}}.} and query the VLM model with \colorbox{blue!10}{\emph{What is this person's occupation?}}. In order to evaluate the model under conditions that closely resemble their usage ``in-the-wild'', we \textit{directly} use the textual output as predicted class and \textit{do not} constrain the output space of the VLM in order to obtain predictions within the label set of the generated dataset.  
More details in Appendix~\ref{app:generationpipeline}.

\textbf{Models.} We again leverage \textsc{Stable Diffusion} \citep{rombach2022high} to generate data in order to investigate models from the \textsc{Blip-2} family with different characteristics (size, instruction tuning). Specifically, we consider \textsc{Blip-2} ViT-g OPT$_{2.7\text{B}}$ (BLIP2-2.7B) with 3.8B parameters and an unsupervised-trained language model, \textsc{Blip-2} ViT-g OPT$_{6.7\text{B}}$ (BLIP2-6.7B), its larger version with 7.8B parameters, and \textsc{Blip-2} ViT-g FlanT5$_{\text{XL}}$ (BLIP2-FlanT5), with 4.1B parameters and an instruction-tuned language model.  

\textbf{Results.} We report effect size for various occupations in Table~\ref{tab:fairness_effectsize} considering predictions by BLIP2-FlanT5.
By comparing the accuracy and \Gap~with effect size for the three classes reported in Table~\ref{fig:motivatingexample}, namely \textit{Writer}, \textit{Doctor}, and \textit{Biologist}, we further validate the main premise of this work. Results for the remaining classes provide evidence that when \Gap~is high, the effect size also increases, further showcasing the potential of such a metric to measure disparities between subgroups that also appear as a mismatch between average and worst-case accuracy.

\begin{table}
\centering
\resizebox{0.43\textwidth}{!}{
\begin{tabular}{c|ccc}
\hline
Occupation      &  Accuracy $(\uparrow)$ & \Gap~$(\downarrow)$ & Effect size $(\downarrow)$ \\ \hline
Writer          & 0.802   & 0.006  & 0.263      \\
Doctor          & 0.903   & 0.073  & 0.291      \\
Biologist       & 0.151   & 0.007  & 0.250      \\ \hdashline
Maid            & 0.317   & 0.120  & 0.556      \\
Model           & 0.838   & 0.102  & 0.368      \\ 
Nurse           & 0.517   & 0.358  & 0.728      \\
Philosopher     & 0.349   & 0.347  & 0.927      \\
Scientist       & 0.737   & 0.065  & 0.241      \\ 
Veterinarian    & 0.791   & 0.001  & 0.154      \\ \hline
\end{tabular}}
\caption{\textbf{VLM evaluation.} Even in cases where the accuracy gap is nearly 0, results show there still is a significant interaction between gender and predicted occupations (e.g. the \textit{Writer} class), indicating the existence of bias that accuracy-based metrics failed to capture.}
\label{tab:fairness_effectsize}
\end{table}

\subsubsection{\metricname~across varying models}

\label{sec:vlm_model_size}
We extend the scope of our evaluation and consider all three instances of the \textsc{Blip2} model family so that we can investigate whether models are biased in different levels, as well as whether specific characteristics such as increased scale and instruction tuning, amplify or mitigate it. 

\begin{figure*}[t]
\centering
\subfigure[Gender bias in occupation prediction. \label{fig:bias_gender_occ}]{\includegraphics[width=0.47\textwidth]{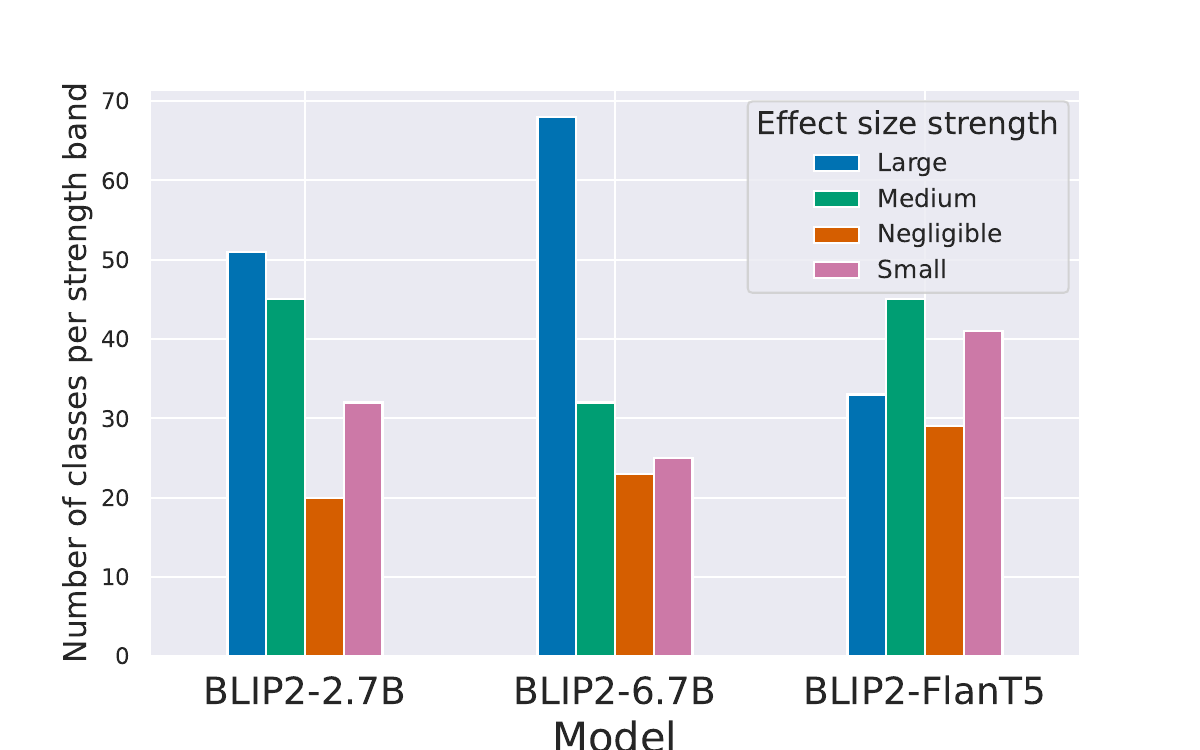}}
\subfigure[Gender bias in sport modality prediction. \label{fig:bias_gender_sports}]{\includegraphics[width=0.47\textwidth]{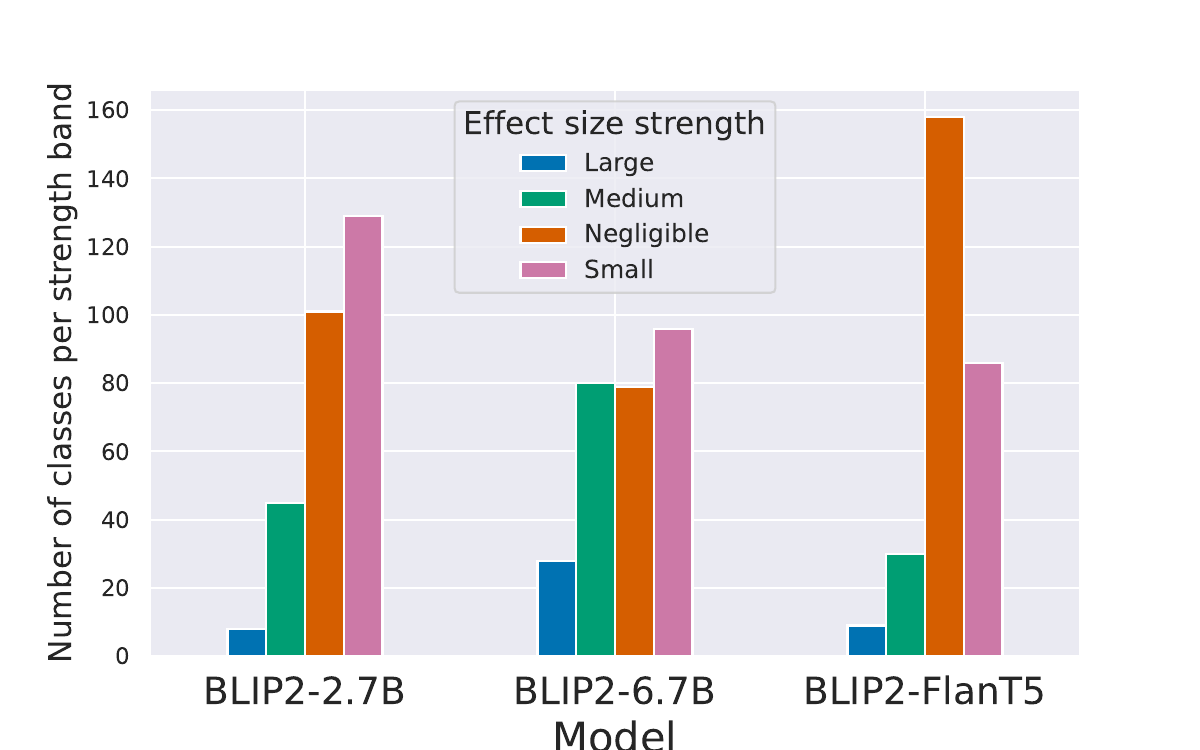}}
\caption{\textbf{Comparing effect size across classes - \textsc{Blip2}.} Splitting effect size values in bands: 0-0.1 is a negligible effect, while 0.1-0.3, 0.3-0.5, and above 0.5 correspond to small, medium, and large, respectively. Scaling up model size with an unsupervised language model increased the amount of large effect size classes, whereas instruction-tuning decreased it.}
\label{fig:bias_gender}
\end{figure*}

\textbf{Effect size strength.} In Figure~\ref{fig:bias_gender}, we categorize effect size values between 0 and 0.1 as negligible\footnote{The use of the word \textit{negligible} here \emph{does not refer} to the extent that potential harms will affect users.} and between 0.1 and 0.3, 0.3 and 0.5, and above 0.5 as small, medium, and large, respectively. For occupation prediction, (Fig.~\ref{fig:dist_v_gender_occ}), larger models have more classes which exhibit medium and large effect sizes, suggesting an overall amplification in gender bias. However, using an instruction-tuned language model leads to fewer classes with large effect in comparison to \textsc{Blip}-2.7B and 6.7B, suggesting instruction tuning may mitigate bias in this instance. Results for sport modality prediction follow a similar trend (Figure~\ref{fig:bias_gender_sports}). The number of classes with negligible effect size decreases when increasing model size, while  \textsc{Blip}2-FlanT5 exhibits less bias.

\textbf{\metricname.} As per  Section~\ref{sec:method}, a fine-grained analysis of the results in Figure~\ref{fig:bias_gender} reveals that the empirical distribution of effect size values (Figures~\ref{fig:dist_v_gender_occ} and~\ref{fig:dist_v_gender_sports}) across all classes is heavy-tailed and left-skewed. We report in Table~\ref{tab:skewness} the skewness coefficients for all models in both tasks. We find that larger models seem to exhibit more bias but instruction fine-tuning seems to mitigate the bias. 

\begin{table}[h]
\begin{center}
\resizebox{0.3\textwidth}{!}{
\begin{tabular}{ccc}
\hline
             & Occupation      & Sports     \\ \hline
BLIP2-2.7B   & 0.233           & 1.205      \\
BLIP2-6.7B   & -0.045          & 0.360      \\
BLIP2-FlanT5 & 0.599           & 1.255      \\ \hline
\end{tabular}}
\caption{\textbf{VLMs \metricname}. Measuring gender bias on occupation and sports modality prediction. Higher \metricname~ values are better. Increasing model size seems to amplify biases, while instruction tuning attenuates it.}
\label{tab:skewness}
\end{center}
\end{table}

\section{Related Work}\label{sec:related_work}

\textbf{Fairness hypothesis testing.}
Previous work has proposed hypothesis testing approaches to probe for fairness under multiple definitions within datasets~\citep{caliskan2017semantics,yik2022identifying} and algorithms~\citep{jourdan2023fairness}. \citet{tramer2017fairtest} introduced a permutation test based on Pearson’s correlation statistic to test for statistical dependence, under a particular metric, between an algorithm's outputs and protected user groups, while \citet{diciccio2020evaluating} proposed to test the hypothesis that a model is fair across two groups as per any given metric. Our work differs from both a methodological perspective, e.g.\ in comparison to \citet{yik2022identifying} which considers whether the data distribution is significantly different from a reference distribution, as well as applicability, since we propose a metric that can capture biases in a multi-class setting, and which goes beyond binary sensitive attributes~\citep{diciccio2020evaluating}. 

\textbf{\bf Evaluating biases in neural network.}
Previous work on bias evaluation has prioritized tasks where the information necessary to measure bias can be directly inferred from text~\citep{rae2021scaling,wang2022measuring,tang2021mitigating,wang2021gender} or by another model \citep{naik2023social}. In contrast, we evaluate bias directly in the model output space, as opposed to relying on predictions of subgroup information. Previous work \citep{birhane2023hate,luccioni2023stable} found  that scale appears to amplify stereotyping and bias, as well as reflect biases in the training data \citep{radford2021learning,wolfe2022american,hall2023vision,prabhu2023lance}.
In the case of VLMs, most prior work focused on leveraging annotated datasets such as MS-COCO~\citep{chen2015microsoft}, CelebA~\citep{liu2015deep} and FairFace~\citep{karkkainen2021fairface} to measure and mitigate bias~\citep{berg2022prompt,chuang2023debiasing,hall2023vision}, while ~\citet{seth2023dear} and ~\citet{smith2023balancing} collected a benchmark and obtained synthetic contrast sets, respectively. Prior work \citep{zhao2017men,wang2021directional} has also evaluated bias amplification, but comparing prediction statistics with the original dataset.

\section{Discussion}\label{sec:conclusions}

In this work, we propose a novel metric, \metricname, to measure biases in classification models, including when the output space is intractable. Motivated by the observation that certain biases may present in the distribution of prediction errors, we draw on tools from contingency table hypothesis testing and propose to measure bias on a per-class basis by estimating the effect size of the interaction between model prediction and the bias variable. Such an approach allows to obtain a scalar value to compare models as well as detailed information about which are the classes mostly affected by biases. \metricname~ does not require any information besides model's outputs to be computed, therefore not introducing any further requirement in comparison to accuracy-based or fairness metrics.  

Experiments across 4 datasets including synthetic examples in cases where no real dataset is available for evaluation, show that \metricname~captures  disparities that accuracy-based metrics do not surface. When the full confusion matrix is available, we also highlight that \metricname~complements standard metrics like demographic parity and equalized odds by identifying classes that are affected by spurious correlations. Our results also show how \metricname~ can be used in practice: the per-class bias profile yielded by our approach sheds light on spurious correlations for classes presenting both higher accuracy and high effect size, while the aggregated \metricname~ is useful to compare models, highlighting, for example, whether increased scale amplifies biases on VLMs, and evaluate the impact of techniques as instruction tuning. 
Our approach can be declined in a variety of previously established metrics, whether based on accuracy, or on fairness criteria \citep{Barocas2019-kj}. In addition, it caters for binary and discrete bias attributes, and could be extended to continuous attributes by using a different statistic \citep[e.g. ][]{tramer2017fairtest,Brown2023}. Aspects to be investigated in future work include employing \metricname~to evaluate previously proposed mitigation strategies for bias in neural networks such as \citep{seth2023dear}, and developing strategies for debiasing.

\textbf{Practical Considerations.} We recommend \metricname~ be employed alongside accuracy-based metrics for a more complete picture of a model's performance. 
Depending on the goal of the evaluation, either per-class measurements can be used or the overall \metricname. In case multiple models are under comparison and the problem involves multiple classes, we recommend using \metricname so that a ranking for the models can be obtained using this overall notion of bias. In cases a specific class is investigated or a fine-grained profile of model bias is required, per-class effect size can be employed. We further note that \metricname~cannot infer a causal relationship between the bias attribute and model predictions, only their association. As with mathematical fairness criteria, our metric does not relate bias to potential harms~\citep{weidinger2022taxonomy}; further work is needed to understand the impact of this distributional bias.
Finally, we remark that is not within the scope of our work to define which biases are practically relevant, given that this is context-dependent and that a metric should account for all existing biases in a dataset/model so that a comprehensive profile of a model's performance can be taken into consideration at the evaluation. 

\clearpage

\section*{Impact Statement}
In this work we propose a metric to estimate how impacted a model is by biases that arise across multiple predictions. We recognize that the binary framing of gender used in the illustrative example and experiments with synthetic data is an oversimplification of an important and complex topic. Our method allows for the interrogation of model bias in terms of discrete, mutually exclusive categories, which may not be ideal for representing multifaceted and intersectional human identities (see \citet{lu2022subverting} for an exploration of this topic). Finally, the synthetic dataset may inherit stereotypes from its generative model, e.g.\ misrepresenting non-cisgender people~\citep{ungless2023stereotypes}.

\section*{Acknowledgements}
We thank Ira Ktena, Lisa Anne Hendricks, Canfer Akbulut, Shakir Mohamed, Simon Osindero, and Florian Stimberg for their feedback throughout the project. We are grateful to Lisa Anne Hendricks and Simon Osindero for their feedback on the manuscript.

\bibliography{bibliography}
\bibliographystyle{icml2024}

\onecolumn
\appendix
\section*{Appendix}
\section{Fairness metrics definitions}\label{sec:appendix_fairness_metrics}

We consider Demographic Parity \citep[DP, ][]{dwork2012} and equalized odds \citep[EO, ][]{hardt2016} as per their multi-class extension described in \citet{alabdulmohsin2022a}. Each class is binarized (one versus all) before the computation of the metric, and the results are aggregated across classes using their maximum value. The metrics refer to a ``maximum gap'' between subgroups. For DP, this would mean computing the proportion of positive predictions for a class in each subgroup, comparing the highest with the lowest values across groups:

\begin{equation}
\mathrm{DP}(f_{\theta}) = \max_{a \in \mathcal Z}\mathbb{E}[f_{\theta}(x) \mid z=a] - \min_{a \in \mathcal Z}\mathbb{E}[f_{\theta}(x) \mid z=a],
\end{equation}

\begin{equation}
    \mathrm{EO}(f_{\theta}) = \max_{a,k \in \mathcal{Z} \times \mathcal Y}\mathbb{E}[f_{\theta}(x) \mid z=a, y=k] - \min_{a,k \in \mathcal{Z} \times \mathcal Y}\mathbb{E}[f_{\theta}(x) \mid z=a, y=k].
\end{equation}

\section{Computing Effect Size using Other Statistics}\label{sec:appendix_phi}
In addition to the 3 accuracy based metrics and 2 fairness metrics we already considered in previous results, in this section we further include the Phi coefficient as a measure of effect size when computing SkewSize in the dSprites experiments. The results in Table \ref{tab:dsprites_phi_coeff} show that in this case the Phi Coefficient yields similar trends as the Cramer’s V. Notice, however, that it is not advisable to use the Phi Coefficient on contingency tables larger than 2x2, which is the reason why we decided to use the more general Cramer’s V when computing \metricname~throughout our work.

\begin{table}[h]
\centering
\begin{tabular}{cccc|ccc}
\hline
                               & \multicolumn{3}{c|}{Cramer’s V} & \multicolumn{3}{c}{Phi Coefficient} \\ \hline
                               & Class 0   & Class 1  & Class 2  & Class 0    & Class 1    & Class 2   \\ \hline
Unbiased                       & 0.012     & 0.011    & 0.019    & 0.012      & 0.011      & 0.027     \\
{\color[HTML]{32CB00} Class 0} & \textbf{0.670}      & 0.015    & 0.016    & \textbf{0.948}      & 0.015      & 0.022     \\
{\color[HTML]{32CB00} Class 1} & 0.014     & \textbf{0.683}    & 0.108    & 0.014      & \textbf{0.966}      & 0.152     \\
{\color[HTML]{32CB00} Class 2} & 0.047     & 0.006    & \textbf{0.696}    & 0.067      & 0.006      & \textbf{0.985}     \\ \hline
\end{tabular}
\caption{Computing effect size with Cram\'er’s V vs Phi Coefficient. \textsc{dSprites} dataset.}
\label{tab:dsprites_phi_coeff}
\end{table}

\section{Estimating distributional bias in multi-class classification: \textsc{DomainNet}}\label{sec:appendix_domainnet}
We now stress-test \metricname~by employing it to evaluate a model in the multi-domain setting, where samples from different distributions are employed training time, and show that our proposed metric can capture systematic biases in predictions. Specifically, we investigate the degree of bias exhibited by the model with respect to the different domains (in this setting, the domain label corresponds to the spurious bias variable).

\textbf{Setting.} We consider the \textsc{DomainNet} benchmark \citep{peng2019moment}, which is composed of images from 6 domains sharing the same label space of 345 object classes, and train a ResNet-50 on the train split of all domains jointly. Given the trained model, we then compute predictions for all instances in the test partitions and proceed to compute \metricname~as per Algorithm~\ref{alg:metric_computation}.

\begin{figure}
  \begin{center}
    \includegraphics[width=0.5\linewidth]{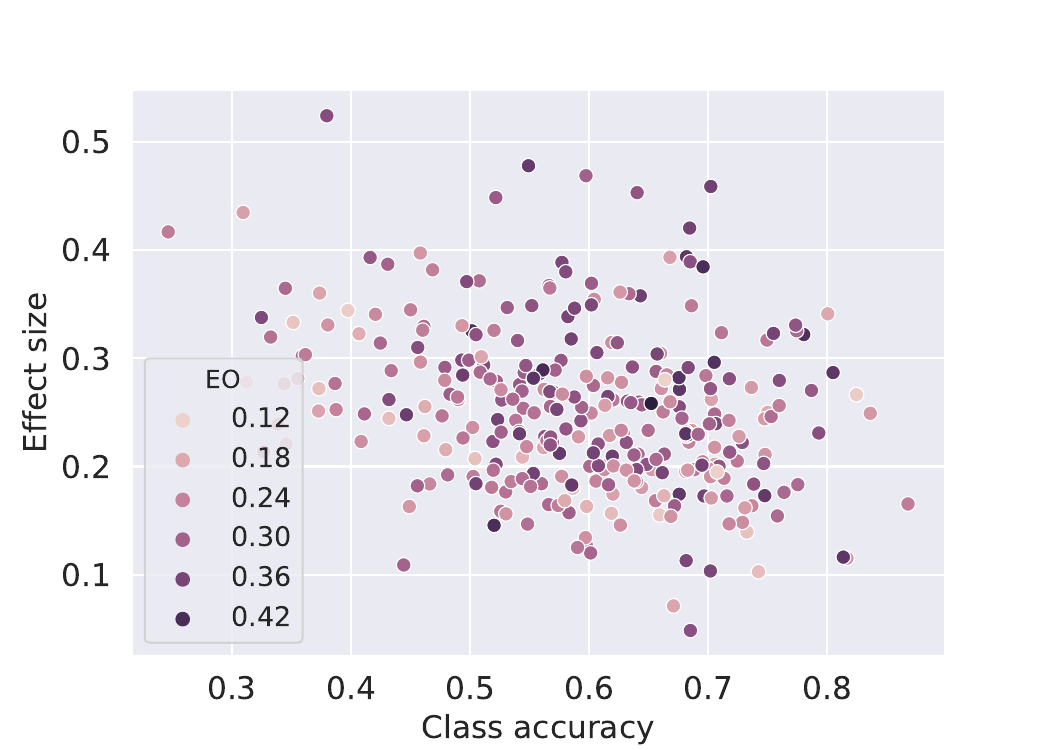}
  \end{center}
  \caption{\textsc{DomainNet.} Per-class accuracy vs. effect size. Hue indicates EO. Points in the top-right most corner of the plot indicate that even for classes where the model is most accurate systematic differences in predictions across subgroups might exist. }
\label{fig:domainnet}
\end{figure}

\textbf{Results.} The model achieved $59.95\%$ average test accuracy, $37.01\%$ worst group accuracy gap, and $0.509$ \metricname. 
In order to provide a fine-grained understanding about the differences between each metric, we show in 
Figure \ref{fig:domainnet} plots accuracy (per class) against effect size $\nu$, along with the respective Equality of Odds (EO) value (shown as each point's corresponding hue). We find a mild Pearson correlation between effect size and accuracy ($-0.291$, $p \approx 0$) as well as between effect size and EO ($0.190$, $p = 0.0008$), which indicate the metrics are related but not equivalent as they capture distinct aspects of the bias. No correlation between effect size and \Gap~was found ($0.103$, $p =0.07$), nor between effect size and DP ($0.051$, $p =0.377$) further highlighting the importance of including robustness evaluations metrics that take into account error mismatches for a given ground-truth class. 

\section{VLM: Detailed results}\label{sec:appendix_results}
\subsection{Optional post-processing}\label{app:post_process}
As we do not constrain the model's output, there may be cases where the model predicts synonyms of the ground-truth class, e.g.\ lawyer and attorney, or the predictions consist of sentences with different structures, e.g.\  \emph{``The person is a laywer''} and \emph{``A lawyer''}. In light of that, in order to compute accuracy values, we manually post-process the outputs of the model to account for all cases where the output semantically matched the ground-truth answer.

\textbf{Impact of post-processing.} We also investigate in Table~\ref{tab:skewness_raw_filt} whether post-processing model outputs affects the overall experimental findings by comparing the metric trend across models for both raw and post-processed outputs.  We find that the same trends can be observed irrespective of the post-processing. Increasing model size while keeping an unsupervised-trained language model amplifies bias as the skewness values decrease when comparing BLIP2-2.7B and BLIP2-6.7B (from 0.233 to -0.045). As expected, \metricname~ values computed with raw model outputs tend to be lower, indicating an overall increase in the computed effect size. This is because, without post-processing, the predicted classes are more fine-grained, resulting in a potential larger mismatch between predictions for each gender value. BLIP2-FlanT5 presented the highest skewness values for all cases, further confirming the findings in Figure~\ref{fig:bias_gender}.

\begin{table}
\centering
\resizebox{0.3\textwidth}{!}{
\begin{tabular}{ccc}
\hline
                              & Raw              & Occupation     \\ \hline
\multirow{2}{*}{BLIP2-2.7B}   & \ding{55} & 0.233   \\
                              & \ding{51} & -0.005  \\ \hline
\multirow{2}{*}{BLIP2-6.7B}   & \ding{55} & -0.045  \\
                              & \ding{51} & -0.130  \\ \hline
\multirow{2}{*}{BLIP2-FlanT5} & \ding{55} & 0.599   \\
                              & \ding{51} & 0.124   \\ \hline
\end{tabular}}
\captionof{table}{\textbf{\metricname for raw versus post-processed model outputs}. Higher skewness values correspond to models having less gender bias. We observe that post-processing the models outputs changes the skewness value but \emph{does not} change the overall trend.}
\label{tab:skewness_raw_filt}
\end{table}

\pagebreak

\subsection{Effect size distributions}
\begin{figure}[h]
\centering
\subfigure[Gender bias in occupation prediction. \label{fig:dist_v_gender_occ}]{\includegraphics[width=0.99\textwidth]{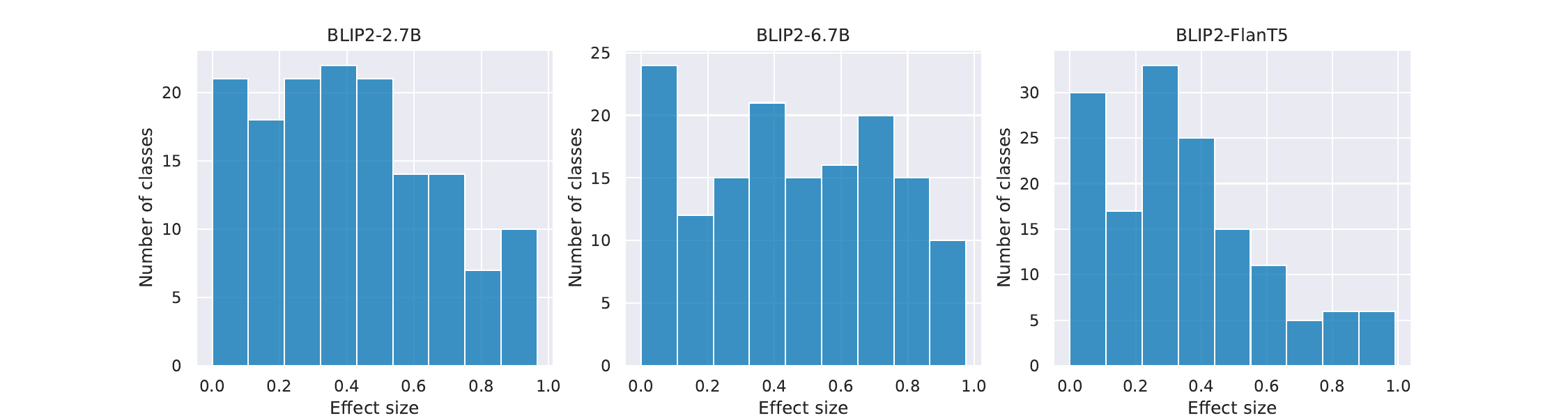}}
\subfigure[Gender bias in sport modality prediction. \label{fig:dist_v_gender_sports}]{\includegraphics[width=0.99\textwidth]{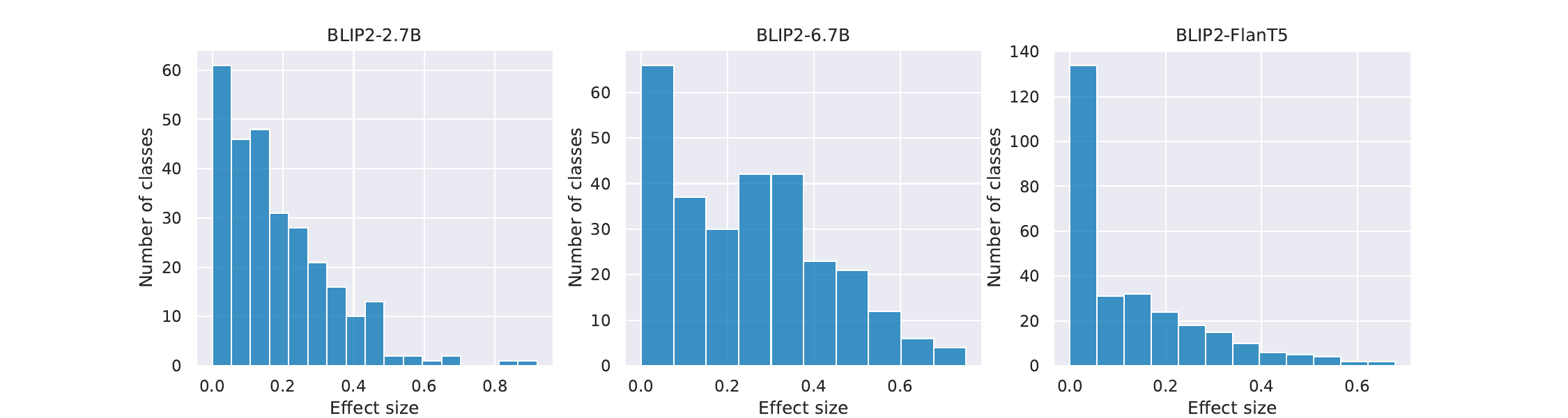}}
\caption{Distribution of effect size values between gender and predicted occupation/sport modality across \textsc{Blip-2} models.}
\label{fig:dist_v_gender}
\end{figure}

\subsection{Controlling the effect of noise in the predictions}\label{sec:appendix_mev}
As the size of output space $\|\mathcal{Y}\|$ grows, we propose the following strategy to control for the sensitivity of \metricname~ to noise in the predictions: as per the rule-of-thumb to satisfy the assumption of the Chi-square test, we can remove columns from the contingency with respective expected value lower than 5.  As we are looking for systematic patterns in the errors of the model, using such a filtering strategy reduces sensitivity to randomness while maintaining sensitivity to the systematic patterns. We can also vary this value in order to decide to which degree some randomness in the predictions should be taken into account.

To illustrate how the choice of the minimum expected value to be accounted for would affect results, we repeated the evaluation reported in Section \ref{sec:vlm_model_size} for the occupation prediction task with varying thresholds so that we can evaluate whether the comparison between models would change. As demonstrated by the results in Table \ref{tab:vlm_mev}, the choice of threshold does not affect the resulting comparison between models.

\begin{table}[h]
\centering
\resizebox{0.5\textwidth}{!}{
\begin{tabular}{cccccc}
\hline
            & MEV=6  & MEV=5  & MEV=4  & MEV=3  & MEV=2  \\ \hline
BLIP-2.7    & 0.235  & 0.233  & 0.225  & 0.199  & 0.19   \\
BLIP-6.7    & -0.031 & -0.045 & -0.056 & -0.072 & -0.102 \\
BLIP-FlanT5 & 0.625  & 0.599  & 0.578  & 0.544  & 0.507  \\ \hline
\end{tabular}}
\captionof{table}{Varying the minimum expected value (MEV) for evaluating the \textsc{Blip2} model family in the occupation prediction task.}
\label{tab:vlm_mev}
\end{table}

\subsection{Data Generation}\label{app:generationpipeline}
We consider 148 and 273 classes for the tasks of occupation and sport modality prediction, respectively. The complete list of occupations and sport modalities used in the VLMs experiments can be found in the Supplementary Material repository.

\section{\textsc{ImageNet} Models}
\label{app:imagenetmodels}
We used a variety of models trained on \textsc{ImageNet} with different sizes, training accuracy, pretraining, etc. Unless otherwise stated, we used publicly available models from \textsc{Tf-Hub}\footnote{https://tfhub.dev/google/imagenet/}.

\begin{itemize}
    \item \textsc{ResNet50-1/2} \citep{he2016deep}: A model we trained on \textsc{ImageNet} from scratch which achieved around $76\%$ accuracy.
    \item \textsc{ResNet*} \citep{he2016deep}: \textsc{ResNet} models with no pretraining.
    \item \textsc{ViT*} \citep{dosovitskiy2020image}: A \textsc{B/16} variant of the vision transformer model family we trained on \textsc{ImageNet} from scratch which achieved around $80\%$ accuracy.
    \item \textsc{Inception*} \citep{szegedy2015going}: Inception models with no pretraining.
    \item \textsc{Inception ResNet} \citep{szegedy2017inception}: A hybrid \textsc{Inception ResNet} model with no pretraining.
    \item \textsc{BiT-S*} \citep{kolesnikov2020big}: \textsc{BiT} models with no pretraining.
\end{itemize}

\section{Further Related Work}

\paragraph{Mitigations.}
Given a known bias in the model, it is possible to mitigate the issue, demonstrating the importance of being able to identify biases to improve the model.
This can be done by intervening on the dataset to make it fairer while maintaining performance as done by \cite{singla2022data}.
Another approach is to intervene on the prompts and de-bias the text embeddings as done by \cite{chuang2023debiasing}.
Finally, we can intervene at the model level, as done by \cite{friedrich2023fair,berg2022prompt} and use guidance or an adversarial loss to steer the model towards being more fair. \citep{zhang2018mitigating}, \citep{alvi2018turning} \citep{kim2023explaining}, \citep{li2022discover}

\paragraph{Datasets for evaluating bias.}
To develop methods to evaluate and mitigate bias, datasets such as Waterbirds \citep{sagawa2019distributionally}, CelebA \citep{liu2015deep}, and MultiNLI \citep{williams2017broad} have been used; in these datasets the biases are created a-priori (e.g.\ land birds on land backgrounds vs water birds on watery backgrounds).
Revise was introduced by \cite{wang2022revise} in order to visualise the biases in a dataset (and thereby probable impacts on models trained on such a dataset).
However, such a tool requires labels on what exists in the dataset, which may not be possible, so \cite{jain2022distilling} demonstrated how biases could be found automatically in multimodal datasets.
A targeted dataset looking at model performance at predicting everyday objects conditioned on geographical location is the DollarStreet dataset \citep{rojas2022dollar}.
Other targeted datasets, such as FairFaces and CasualConversationsV2 \citep{karkkainen2021fairface,porgali2023casual,hall2023visogender} can be used to evaluate a models bias across sensitive attributes. \cite{karkkainen2021fairface,porgali2023casual} does so by comparing classification performance across these attributes (e.g.\ age, gender, etc.) whereas \cite{hall2023visogender} is a small dataset of ~250 images that evaluates pronoun resolution conditioned on the image. Such datasets are complementary to our approach, as we rely on a dataset with labelled subgroups to compute our metric. When no such dataset exists, we relied on synthetic data.

\newpage
\section{\metricname~ implementation details}
\subsection{Pseudocode}
\begin{algorithm}
	\caption{Computing \metricname} 
	\begin{algorithmic}[1]
		\FOR{$i=1,2,\ldots,|\mathcal{Y}|$}
		    \STATE Get set of model predictions $\hat{Y}^i=\{\hat{y}_k\}$ for all $(x_k, y_k, z_k)$ where $y_k=y_i$
			\FOR{$j=1,2,\ldots,|\mathcal{Z}|$}
				\STATE Build $\hat{Y}^{ij}$, a subset of $\hat{Y}^i$ with instances where $z_k=z_j$  
			\ENDFOR
			\STATE Estimate $\nu_i$, the effect size for the $i$-th class, using Equation~\ref{eq:effect_size}.
		\ENDFOR
		\STATE Aggregate effect size estimates per class by computing \metricname~as per Equation~\ref{eq:metric_def}.
	\end{algorithmic}
	\label{alg:metric_computation}
\end{algorithm}

\subsection{Python Implementation}
\begin{python}
# Copyright 2023 The SkewSize Authors. All rights reserved.
# SPDX-License-Identifier: Apache-2.0

import numpy as np
import pandas as pd
import scipy.stats as stats

v_list = []
for label in unique_labels:
    # predictions: predictions for all instances in the class *label*.
    # subgroups: predictions for all instances in the class *label*.
    df = pd.DataFrame({'predictions': predictions,
                       'subgroups': subgroups})
    crosstab = pd.crosstab(df.subgroups, df.predictions)

    chi2 = stats.chi2_contingency(crosstab)[0]
    dof = min(crosstab.shape)-1
    n = crosstab.sum().sum()
    v = np.sqrt(chi2/(n*dof))
    v_list.append(v)
    
v_values = np.asarray(v_list)
# When a model predicts correctly all examples 
# in a given class across all subgroups
# dof=0 and the corresponding v is NaN. 
# We remove NaNs before computing skewsize.
v_values = v_values[~np.isnan(v_values)]
skewsize = stats.skew(v_values)

\end{python}

\end{document}